\documentclass[journal]{IEEEtran}

\usepackage{times}
\usepackage{epsfig}
\usepackage{graphicx}
\usepackage{amsmath}
\usepackage{amssymb}

\usepackage{multirow}
\usepackage{algorithm}
\usepackage[ruled,vlined,algo2e]{algorithm2e}
\usepackage[table,xcdraw]{xcolor}

\usepackage{graphicx}
\usepackage{textcomp}
\usepackage{booktabs}
\usepackage{multirow}
\usepackage{diagbox}
\usepackage{float}
\usepackage{subfigure}
\usepackage{enumerate}
\usepackage[colorlinks,linkcolor=black]{hyperref}
\usepackage{threeparttable}
\usepackage{amsmath,amssymb,amsfonts}
\usepackage[ruled]{algorithm2e}
\usepackage{algpseudocode}
\usepackage{colortbl} 
\usepackage{url}
\usepackage{bbding}
\usepackage{amsmath}
\definecolor{mycyan}{cmyk}{.3,0,0,0}
\definecolor{mypink}{rgb}{.99,.91,.95}
\definecolor{mygray}{gray}{.9}

\usepackage{cite}
\usepackage{hyperref}

\begin{document}

\title{Graph Relation Distillation for Efficient Biomedical Instance Segmentation}

\author{Xiaoyu~Liu,
 Yueyi Zhang, Zhiwei Xiong, Wei Huang, Bo Hu, Xiaoyan Sun and Feng Wu

\thanks{The authors are with the University of Science and Technology of China and the Institute of Artificial Intelligence, Hefei Comprehensive National Science Center (e-mail: \{liuxyu, weih527, hubosist\}@mail.ustc.edu.cn; \{zhyuey, zwxiong, sunxiaoyan, fengwu\}@ustc.edu.cn).}
}

\markboth{ }%
{Liu \MakeLowercase{\textit{et al.}}: Graph Relation Distillation for Efficient Biomedical Instance Segmentation}
%

\maketitle
\makeatletter
\def\ps@IEEEtitlepagestyle{
  \def\@oddfoot{\mycopyrightnotice}
  \def\@evenfoot{}
}
\def\mycopyrightnotice{
  {\footnotesize
  \begin{minipage}{\textwidth}
  \centering
  
  \end{minipage}
  }
}


\begin{abstract} 
    Instance-aware embeddings predicted by deep neural networks have revolutionized biomedical instance segmentation, but its resource requirements are substantial. Knowledge distillation offers a solution by transferring distilled knowledge from heavy teacher networks to lightweight yet high-performance student networks. However, existing knowledge distillation methods struggle to extract knowledge for distinguishing instances and overlook global relation information. To address these challenges, we propose a graph relation distillation approach for efficient biomedical instance segmentation, which considers three essential types of knowledge: instance-level features, instance relations, and pixel-level boundaries. We introduce two graph distillation schemes deployed at both the intra-image level and the inter-image level: instance graph distillation (IGD) and affinity graph distillation (AGD). IGD constructs a graph representing instance features and relations, transferring these two types of knowledge by enforcing instance graph consistency. AGD constructs an affinity graph representing pixel relations to capture structured knowledge of instance boundaries, transferring boundary-related knowledge by ensuring pixel affinity consistency. Experimental results on a number of biomedical datasets validate the effectiveness of our approach, enabling student models with less than $ 1\%$ parameters and less than $10\%$ inference time while achieving promising performance compared to teacher models. The code is available at \href{https://github.com/liuxy1103/GIDBIS}{https://github.com/liuxy1103/GRDBIS}.
    
 \end{abstract}


\ifCLASSOPTIONpeerreview
\begin{center} \bfseries EDICS Category: 3-BBND \end{center}
\fi
%
\IEEEpeerreviewmaketitle



\section{Introduction}
\label{sec1}

\IEEEPARstart{B}{iomedical} instance segmentation is a crucial and highly challenging task in the field of biomedical image analysis. Its objective is to assign a unique identification to each instance in the entire image.
The two most common approaches are based on either performing semantic segmentation of instance contours~\cite{chen2016dcan,li2022advanced,kumar2017dataset,song2023muscleparsenet} or object detection approaches~\cite{he2017mask,liu2019nuclei,zhang2018panoptic}. These methods often struggle when faced with unclear boundaries, densely distributed instances, and significant occlusions.
Recent advancements ~\cite{chen2019instance,kulikov2020instance,lee2021learning,payer2018instance} in this field employ convolutional neural networks (CNNs) to predict instance-aware embedding vectors that are unaffected by object morphology. Subsequently, post-processing algorithms are employed to cluster pixel embeddings into instances. Nonetheless, these methods heavily rely on complex models that demand substantial computational resources to capture detailed instance-aware features for pixel-level dense estimation. Such resource-intensive models are impractical for real-world scenarios. Therefore, the development of simplified networks is of paramount importance, particularly in the case of 3D CNNs. Consequently, there exists a trade-off between model simplification for speed and maintaining optimal performance, as the simplified models often compromise their performance.


Knowledge distillation has emerged as a highly promising approach to reduce computational costs while maintaining satisfactory performance~\cite{hinton2015distilling,romero2014fitnets,tung2019similarity,zagoruyko2016paying}. Through knowledge distillation, it becomes possible to train a lightweight student network that achieves high performance by leveraging effective knowledge distilled from a well-trained and computationally intensive teacher network. Recently, knowledge distillation methods have been primarily designed for image-level classification and semantic segmentation tasks. For instance, feature distillation has been used in ~\cite{zagoruyko2016paying}, which transfers attention maps distilled from mid-level feature maps of the teacher network to the student network. Logit distillation proposed by ~\cite{qin2021efficient} introduces a technique that distills both the output logits and semantic region information from the teacher network to guide the student network. These advancements demonstrate the potential of knowledge distillation to enhance various aspects of model training and transfer valuable insights from heavy models to lightweight counterparts.


However, applying existing knowledge distillation methods directly to instance segmentation encounters two main challenges. Firstly, instance-level features and the relations between instances play a crucial role in distinguishing between different instances based on their feature space distances. Unfortunately, current methods often overlook this vital information. Unlike the straightforward logit distillation employed in semantic segmentation, instance segmentation demands more sophisticated techniques to distill structural information of instance boundaries from feature maps.
While a general distillation method  in~\cite{chen2021distilling} is proposed for multiple vision tasks, which incorporates a review mechanism involving multiple feature maps, this approach encounters difficulties in learning knowledge that contains redundant information. This is primarily due to the limited ability of lightweight networks to adequately attend to features at each pixel location. As a result, there has been a rare exploration of knowledge distillation methods tailored to address the unique challenges posed by biomedical instance segmentation.

Secondly, existing knowledge distillation methods primarily focus on extracting valuable knowledge based on individual input images to guide a student network. Unfortunately, they tend to neglect the essential inter-image instance relations at both pixel-level and instance-level, hindering effective knowledge transfer. It is important to highlight that global relations across different input images encompass valuable instance structural information. Incorporating these global relations becomes instrumental in constructing a well-structured feature space and attaining more precise instance segmentation results.

In this paper, we propose a novel graph relation distillation method tailored for biomedical instance segmentation, which addresses the challenges faced by existing distillation methods. To tackle the first challenge, we introduce two distillation schemes to extract crucial knowledge for instance segmentation, including instance-level features, instance relations, and pixel-level boundaries. The first scheme, instance graph distillation (IGD), constructs an instance graph using central embeddings of the corresponding instances as nodes and measuring feature similarity between nodes as edges. By enforcing graph consistency, IGD effectively transfers knowledge of instance features and relations. The second scheme, affinity graph distillation (AGD), regards each pixel embedding as a graph node and converts pixel embeddings into a structured affinity graph by calculating the distance between pixel embeddings, which is used to mitigate boundary ambiguity in the lightweight student network. AGD ensures affinity graph consistency between teacher and student networks, facilitating boundary-related knowledge transfer. By employing the above two distillation schemes, our approach enhances the performance of student networks in biomedical instance segmentation. 

To address the second challenge, we extend the IGD and AGD schemes to capture global structural information at the inter-image level. Specifically, we construct instance graphs and affinity graphs by considering relations between instances and pixel embeddings from different input images, respectively. To fully explore the graph relations between different input images, we need to increase the batch size of the network by including as many input samples as possible. Under the constraint of limited GPU memory, we introduce a memory bank mechanism to store past predicted feature maps as much as possible. This enables us to calculate relations between the current input image and sampled images from the memory bank, effectively capturing long-range inter-image relations. Overall, our approach offers a practical solution for distilling knowledge vital to biomedical instance segmentation, addressing limitations and improving performance.

Extensive experimental results show that our knowledge distillation approach greatly benefits the lightweight network, leading to significant improvements in performance while maintaining efficiency during inference. The student networks trained using our approach achieve promising performance with less than 1\% of parameters and less than 10\% of inference time compared to the teacher networks.

The contributions of this paper are summarized as follows: 
\begin{itemize}

\item We propose a graph relation distillation method tailored for biomedical instance segmentation to obtain efficient and high-performance networks.



\item We propose an IGD scheme to force the student network to mimic instance-level features and instance relations of the teacher network via an instance graph, along with an AGD scheme for pixel-level boundary knowledge transfer.

\item We extend the IGD and AGD schemes from the intra-image level to the inter-image level by introducing a memory bank mechanism to capture the global relation across different input images.

\item The superiority of our knowledge distillation approach over existing methods is demonstrated on three 2D biomedical datasets and two 3D biomedical datasets. 

\end{itemize}

This work is a substantial extension of our preliminary work~\cite{liu2022efficient} in the following aspects:
\begin{itemize}
\item We distill global relation information across different input images using an inter-image instance graph and inter-image affinity. This enables the student network to learn from a broader range of instance features and boundary structures and improve its ability to handle complex instances.

\item We verify the effectiveness of our proposed method for more teacher-student network pairs with different architectures, demonstrating its versatility and robustness.

\item We conduct extensive experiments on biomedical instance segmentation datasets with different modalities and multiple categories, further demonstrating the effectiveness and generalizability of our approach. 
\end{itemize}

\section{Related Works}

\begin{figure*}[!htb]
\includegraphics[width= \textwidth]{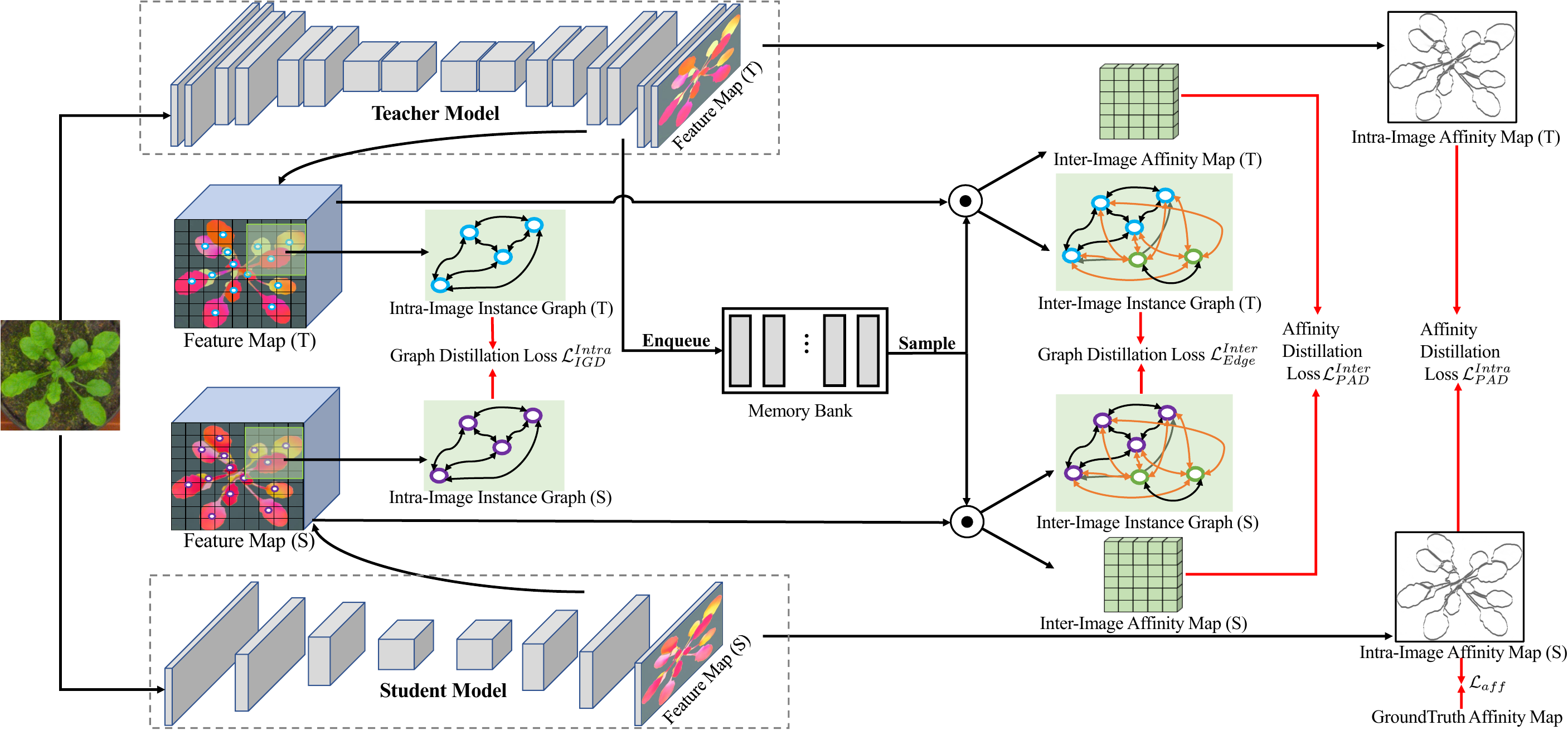}
\caption{Workflow of our proposed graph relation distillation method for biomedical instance segmentation, which includes two schemes. The instance graph distillation (IGD) scheme constructs instance graphs from embeddings of the teacher-student network pair and enforces the consistency of graphs constructed by the teacher, while the affinity graph distillation (AGD) scheme converts pixel embeddings into pixel affinities that encode structured information of instance boundaries and enforces the student model to generate affinities similar to its teacher model. These two schemes take charge of the knowledge distillation mechanism and are carried out at both intra-image and inter-image levels for global instance relations. The symbol $\odot$ represents dot product operation. The red arrow indicates the loss function.}
\label{fig1}
\end{figure*}

\subsection{Biomedical Instance Segmentation}
Deep learning-based instance segmentation methods for biomedical images can be classified into two main categories: proposal-free and proposal-based methods. 

Proposal-based methods \cite{
he2017mask,liu2019nuclei,zhang2018panoptic,liu2021panoptic,kirillov2019panoptic,yi2019attentive} utilize object detection~\cite{girshick2015fast,zhang2021segmenting,zhang2019mask,yang2021task} and object segmentation heads to predict bounding boxes and foreground masks for each object, respectively. However, these methods heavily rely on accurate bounding box predictions, which may fail to differentiate adjacent instances due to their overlap, and the size of instances. The sizes of instances in images may exceed the receptive field of the model, making it challenging to locate complete instances using bounding boxes.

The proposal-free methods~\cite{de2017semantic,payer2018instance,kulikov2020instance,lalit2022embedseg} predict specially designed instance-aware features and morphology properties which can encode morphology characteristics~\cite{shi2022polyp}, structures, and spatial arrangement, and then cluster the predicted mid-representations into instances using a post-processing algorithm~\cite{beier2017multicut,fukunaga1975estimation,comaniciu2002mean,campello2013density}. 
Pixel embedding-based methods ~\cite{wolny2022sparse,lalit2021embedding,payer2019segmenting,huang2022learning,liu2021biological} excel in encoding each pixel of an image into a high-dimensional feature space, facilitating the grouping of similar pixels to form distinct instance regions. These methods exhibit exceptional performance in tackling complex scenes with overlapping and crowded objects, making them a popular choice for biomedical instance segmentation and other applications.


However, the computational and memory demands of pixel embedding-based methods hinder their practical use. Knowledge distillation offers a solution by transferring knowledge from a large, complex model (the teacher) to a smaller, simpler model (the student). This compression process effectively preserves the teacher network's knowledge in the student network, maintaining comparable performance. By applying knowledge distillation to pixel embedding-based segmentation models, we can significantly reduce their computational and storage requirements while preserving their exceptional segmentation performance.

\subsection{Knowledge Distillation}
The goal of knowledge distillation is to transfer knowledge from a computationally expensive but powerful teacher network to a lightweight student network, thereby enhancing its performance while preserving its efficiency. Related work on knowledge distillation can be broadly classified into the following several categories, depending on the specific methods used for distillation.

One approach is to distill knowledge from a larger teacher network to a smaller student network. This can be done by directly transferring the output probabilities of the teacher network to the student network~\cite{hinton2015distilling}, or by transferring intermediate representations from the teacher network to the student network~\cite{romero2014fitnets}. Other methods use attention mechanisms or gating functions to allow the student network to selectively focus on the most important information from the teacher network~\cite{zagoruyko2016paying}.
Another approach is to use ensemble models as teachers to transfer knowledge to a single student network. This can be done by using the logits or probabilities output by the ensemble as a soft label for the student network~\cite{fukuda2017efficient,chao2021rethinking}, or by transferring feature maps or attention maps from the ensemble to the student network~\cite{noothout2022knowledge}.
In addition to these methods for general tasks, there are also methods for task-specific knowledge distillation, such as for image classification~\cite{liu2019knowledge,li2023boosting,xu2023improving}, semantic segmentation~\cite{qin2021efficient,wen2021towards}, object detection~\cite{chen2021deep}, and speech recognition~\cite{fu2019paraphrase}.

However, there is currently no proposed knowledge distillation method tailored for biomedical instance segmentation, which is a challenging task due to the complexity and heterogeneity of biomedical images with instances varying significantly in size, shape, and distribution. Therefore, the proposed method is a novel application of knowledge distillation to address the challenges of biomedical instance segmentation.
Existing works~\cite{chen2021deep,liu2019knowledge} consider relation distillation for image classification and object detection by constructing a graph.
In this paper, we extend this idea to biomedical instance segmentation by constructing a graph from predicted pixel embeddings and considering cross-image relations with corresponding domain knowledge. Furthermore, we extend this concept to biomedical instance segmentation. Specifically, we construct an instance graph and an affinity graph based on the predicted pixel embeddings and incorporate cross-image relations, by leveraging relevant domain knowledge. By doing so, we aim to improve the performance of biomedical instance segmentation and enhance the understanding of inter-instance relations and pixel relations within the biomedical domain.

\section{Methodology}



The workflow of our proposed distillation method is presented in Fig. \ref{fig1} and can be applied to both 2D and 3D networks for images and volumes. We illustrate it using a 2D image example for easy visualization and description. The method involves a heavy teacher network $T$ and a lightweight student network $S$, which both predict a set of feature maps, \textit{i.e.}, embedding map $E\in \mathbb{R}^{D \times H\times W}$ for an input image of size $H\times W$. The embedding vector of a pixel $p$ is denoted as $e_p\in \mathbb{R}^{D}$ and can be clustered into instances through post-processing, and $D$ is the dimension of the embedding vectors. Given $M$ training images, the segmentation network can extract $M$ embedding maps $\left\{\mathbf{E}_{m} \in \mathbb{R}^{D \times H \times W}\right\}_{m=1}^{M}$. Two specially designed schemes, instance graph distillation (IGD) and affinity graph distillation (AGD), are employed to distill effective knowledge from embedding maps at both intra-image and inter-image levels. More details of our proposed method are provided below.

\subsection{Instance Graph Distillation}

Embeddings of pixels $p \in \mathbb{S}^{i}$ belonging to the same instance $i$ and located in the corresponding area $\mathbb{S}^i$ exhibit similarity, while the embeddings of pixels belonging to different instances demonstrate dissimilarity. This ensures that the different instances $i \in \mathbb{I}$ of an input image are distinguished in the feature space. Therefore, the distribution of embeddings in the feature space contains valuable knowledge related to instance-level features and instance relations. To transfer this key knowledge, we propose an instance graph distillation scheme.

\subsubsection{Intra-Image Distillation}

To effectively distill this knowledge at the intra-image level, we construct an intra-image instance graph that encodes the knowledge of instance-level features and instance relations by nodes and edges, respectively. The nodes are extracted from pixel embeddings of the embedding map with the guidance of labeled instance masks which provide precise areas to calculate instance central features, denoted as
\begin{equation}
\begin{aligned}
v_i=\frac{1}{|\mathbb{S}^i|}\sum_{p \in \mathbb{S}^i}^{}e_p.
\end{aligned}
\end{equation}

The edges are defined as the cosine distance between two nodes, denoted as
\begin{equation}
\begin{aligned}
\varepsilon_{ij}=Cos(v_i,v_j)=\frac{v_i^{\top}\cdot v_j}{\left \| v_i \right \| \left \| v_j \right \| }, 
\end{aligned}
\end{equation}
where instances $i,j\in \mathbb{I}$, $i\ne j$, and $\top$ represents vector transpose operation, and $Cos$ is a function to calculate cosine similarity distance between two vectors.

We then force the instance graph of the student network to be consistent with the instance graph of the teacher network. The distillation loss of this scheme $\mathcal{L}_{IGD}^{intra}$ can be divided into two parts, respectively related to nodes and edges:
\begin{equation}
\begin{aligned}
\mathcal{L}_{IGD}^{Intra}&= \lambda_1\mathcal{L}_{Node}^{Intra}+\lambda_2\mathcal{L}_{Edge}^{Intra}, \\
\mathcal{L}_{Node}^{Intra} &=\frac{1}{\left | \mathbb{I}  \right | }\sum_{i\in\mathbb{I} }^{} \left \| (v_i)^S-(v_i)^T \right \|_2,\\
\mathcal{L}_{Edge}^{Intra} &= \frac{1}{\left | \mathbb{I}  \right |^2}\sum_{i\in\mathbb{I} }^{} \sum_{j\in\mathbb{I} }^{}\left \| (\varepsilon_{ij})^S-(\varepsilon_{ij})^T \right \|_2,
\end{aligned}
\end{equation}
where $\lambda_1$ and $\lambda_2$ are weighting coefficients to balance the two terms, and the superscripts $T$ and $S$ represent the teacher and student networks.




\subsubsection{Inter-Image Distillation}
With the goal of transferring the instance relations across different input images, we extend the above-mentioned instance graph distillation scheme to the inter-image level. Given the limitation of GPU memory, we follow \cite{tian2020contrastive, yang2022cross} to introduce a shared online feature map queue between the student and teacher networks, which stores a vast quantity of feature maps in a memory bank generated from the predictions of the teacher network in previous iterations. It allows us to retrieve abundant feature maps efficiently.
The feature map queue can store $K$ feature maps, and is notated as $\left\{\mathbf{E}_{k} \in \mathbb{R}^{H \times W \times d}\right\}_{k=1}^{K}$.
In each training iteration, we enqueue batch-size $B$ feature maps to the memory bank and randomly sample $L$ feature maps from it.
Given the $m^{th}$ input image from the training image, the segmentation network can predict the embedding map $E_m \in \mathbb{R}^{D \times H\times W}$. Meanwhile, we can obtain $L$ embedding maps $\left\{\mathbf{E}_{l} \in \mathbb{R}^{D \times H \times W }\right\}_{l=1}^{L}$ sampled from the memory bank.

We then calculate the corresponding node features $v_i^m$ and $v_j^l$ extracted from these two feature maps $E_m$ and  $E_l$, where $i \in \mathbb{I}^m$ and $j \in \mathbb{I}^l$ represent different instances from the $m^{th}$ input image and the $l^{th}$ sampled image from the memory bank, respectively. The edge feature between two nodes $v_i^m$ and $v_j^l$ is calculated as $\varepsilon_{ij}^{ml}$ by the above-mentioned cosine distance.

Given that the relations between instances within an image have been leveraged by intra-image distillation, we build the inter-image instance graph by only connecting nodes from different input images. We enforce consistency between two inter-image graphs respectively constructed from the student network and the teacher network, by using an MSE loss function. It is formulated as follows:
\begin{equation}
\begin{aligned}
\mathcal{L}_{Edge}^{Inter}=\frac{1}{L \left | \mathbb{I}^m  \right |\left | \mathbb{I}^l  \right |}\sum_{l=1 }^{L}\sum_{i\in\mathbb{I}^m }^{} \sum_{j\in\mathbb{I}^l }^{}\left \| (\varepsilon_{ij}^{ml})^S-(\varepsilon_{ij}^{ml})^T \right \|_2.
\end{aligned}
\end{equation}

 


\subsection{Affinity Graph Distillation}

To transfer the structured information of instance boundaries, we further propose an AGD scheme to convert pixel embeddings into an affinity graph that encodes pixel relations and calculate the mean square error loss between two affinity graphs respectively from the teacher and student networks. 

\subsubsection{Intra-Image Distillation}
The pixel affinity, as a node in the affinity graph, describes the relation between pixel embeddings. We first compute the pixel affinity at the intra-image level, which is formulated as $a_{n,p} = Cos(e_p, e_{p+n})$, 
where pixel $p$ and pixel $p+n$ are constrained to be locally adjacent within $n$ pixel strides to ensure efficient use of the embedding space, as demonstrated in~\cite{chen2019instance}. Accordingly, a feature map $E$ is converted into the affinity map $A \in \mathbb{R}^{N \times H\times W}$, where $N$ represents the adjacent relations within $N$ different pixel strides.

To distill instance structure information at the intra-image level from the teacher network to the student network, we align the affinity maps generated by the teacher and student networks. It can improve the ability of students to capture the structure of instance boundaries.  We denote the teacher and student affinity maps as $A^T$ and $A^S$, respectively, and then force $A^S$ to be consistent with $A^T$:
\begin{equation}
\begin{aligned}
\mathcal{L}_{AGD}^{Intra} &= \left \| A^S-A^T \right \|_2  \\
        &= \frac{1}{N\times H\times W}\sum_{n=1}^{N}\sum_{p=1}^{H\times W}\left \| a_{n,p}^S-a_{n,p}^T \right \|_2.
\end{aligned}
\end{equation}

The affinity map converted from the embedding map of the last layer of the student network is also supervised by the affinity label $\hat{A}$ from the ground-truth segmentation, we formulate the loss as
\begin{equation}
\begin{aligned}
\mathcal{L}_{aff} = \left \| A^S - \hat{A} \right \|_2.
\end{aligned}
\end{equation}

\subsubsection{Inter-Image Distillation}
To distill global instance structure information across different input images, we define inter-image affinity as the similarity between pixel embeddings from different input images. By comparing the pixel embeddings from different input images, we can infer the similarities and differences between the instances present in each image. This allows the student network to extract and analyze the global instance structure information.  Similar to the inter-image graph distillation, We calculate the inter-image affinity between pixel embeddings $e_i^m \in E_m$ and $e_j^l \in E_l$ as $a_{ij}^{ml} = Cos(e_i^m,e_j^l)$.

The inter-image affinity map, denoted as $A^{ml} \in \mathbb{R}^{HW \times HW}$, is calculated by taking the dot product of the pixel embeddings of two images, \textit{i.e.}, $A^{ml} = E_m^{\top}E_l$. This calculation captures inter-image pair-wise relations among all pixel embeddings. We guide the inter-image affinity map $(A^{ml})^S$ produced from the student network to be consistent with $(A^{ml})^T$ from the teacher network. It is formulated as
\begin{equation}
\begin{aligned}
\mathcal{L}_{AGD}^{Inter} &= \frac{1}{L}\sum_{l=1}^{L}\left \| (A^{ml})^S-(A^{ml})^T \right \|_2  \\
        &= \frac{1}{L\times(H\times W)^2}\sum_{l=1}^{L}\sum_{i,j=1}^{H\times W}\left \| (a_{ij}^{ml})^S-(a_{ij}^{ml})^T \right \|_2.
\end{aligned}
\end{equation}

\subsection{Overall Optimization}
Given a well-trained teacher network, the objective function integrates all knowledge distillation schemes mentioned above for training the student network. The total loss is formulated as

\begin{equation}
\begin{aligned}
\mathcal{L}_{total} &= \mathcal{L}_{aff}+\underbrace{\lambda_1\mathcal{L}_{Node}^{Intra} +\lambda_2\mathcal{L}_{Edge}^{Intra}}_{\mathcal{L}_{IGD}^{Intra}}\\
&+\lambda_3\mathcal{L}_{AGD}^{Intra} +\lambda_4\mathcal{L}_{Edge}^{Inter}+\lambda_5\mathcal{L}_{AGD}^{Inter},
\end{aligned}
\end{equation}
where $\lambda_1$, $\lambda_2$, $\lambda_3$, $\lambda_4$, and $\lambda_5$ are empirically set as 0.1, 0.1, 10, 1, and 1 through experiments to balance these terms. During the inference phase, we employ standard post-processing algorithms to generate the ultimate instance-level segmentation results from the predicted embedding maps, following~\cite{huang2022learning}. For 3D networks, we utilize well-established post-processing algorithms such as Waterz~\cite{funke2018large} and LMC~\cite{beier2017multicut}. For 2D networks, we rely on the Mutex algorithm~\cite{wolf2020mutex} to generate the segmentation results. 



\subsection{Network Structure}

\subsubsection{Teacher networks}

As shown in Table~\ref{setting}, we provide an overview of the model complexity and inference time on test datasets for different teacher and student networks. We adopt two state-of-the-art heavy networks as teachers, namely 3D U-Net MALA~\cite{funke2018large} for 3D datasets, and 2D ResUnet~\cite{anas2017clinical} and NestedUNet~\cite{zhou2018unet++} for 2D datasets, to demonstrate the effectiveness of our knowledge distillation method.

U-Net MALA is a modified 3D version of U-Net~\cite{ronneberger2015u} for 3D EM image segmentation. It has 4 levels with at least 1 convolution pass per level, using max pooling for downsampling and transposed convolution for upsampling. The resulting maps are concatenated with feature maps from the downsampling pass of the same level, and cropped to account for context loss. ResUnet and NestedUNet are both variants of U-Net, which have impressive performance on biomedical image segmentation tasks. ResUnet uses residual blocks to avoid vanishing gradient, while NestedUNet has multiple nested skip connections for multi-level feature access.


\begin{table}[!t]\fontsize{8.9}{10}\selectfont
\centering
\caption{Model complexity and inference time on test datasets for different teacher and student networks. The FLOPS and inference time are estimated with the input size of $544\times544$ on the CVPPP dataset for 2D backbones and with the input size of $84\times268\times268$ on the Cremi dataset for 3D UNet-MALA backbones. The inference time is calculated as the sum of all input samples in the entire test set. The symbols `T' and `S' represent the teacher network and the student network, respectively.}\label{setting}
        \renewcommand\tabcolsep{3.5pt}
	\begin{center}
		\begin{tabular}{l|c|c|c}
 			\midrule
			Models & \#Params (M) & FLOPs (GMAC) & \#Infer Time (s)\\
			\midrule
   			T: ResUNet &33.61 & 582.97  & 2.0 $\pm$ 0.1 \\
			T: NestedUNet &36.36 & 624.14 & 2.1 $\pm$ 0.1 \\
                T: MALA & 84.02 &  367.04 & 53.5 $\pm$ 2.0 \\
                \cmidrule(r){1-4}
			S: ResUNet-tiny &0.30 & 5.76 & 0.20 $\pm$ 0.03\\
         	S: MobileNet &4.79 & 25.82 & 0.78  $\pm$ 0.05\\
                S: MALA-tiny  & 0.37& 22.01 & 2.7 $\pm$ 0.2\\
            \bottomrule
		\end{tabular}
	\end{center}
\end{table}

\subsubsection{Student networks}
To fully validate the effectiveness and versatility of the proposed distillation methods, we conduct experiments for two kinds of student networks, which are described as follows.
1) The dimension of pixel embeddings plays a crucial role in capturing instance features accurately. Recognizing that the high-dimensional feature maps in the teacher network contain redundancy, we adopt a straightforward approach. We proportionally reduce the width of each layer in the teacher networks to match their corresponding lightweight student networks. This allows the student networks to maintain the same architecture as the teacher networks, facilitating the learning process as they absorb the distilled knowledge from their teachers. As shown in Table~\ref{setting}, the student networks `MALA-tiny' and `ResUNet-tiny' have reduced the width of each layer in the two teacher networks to approximately $\frac{1}{10}$ and $\frac{1}{5}$ of their original width, respectively. These two student networks have only 0.4\% and 0.9\% of the parameters of their corresponding teacher networks, `MALA' and `ResUNet', and consume only 5\% and 10\% of the inference time required by their teacher networks. We also conduct more detailed ablation experiments in Ablation Study \ref{ablation_models} to explore the distillation performance of a series of small models obtained by reducing the width of each network layer in different ratios.

2) We employ the well-established lightweight network MobileNetV2~\cite{sandler2018mobilenetv2} as the student network. MobileNetV2 utilizes depth-wise and point-wise convolutions, resulting in reduced parameters and computation compared to traditional convolutions. This network architecture is widely adopted in mobile and embedded vision applications. It is worth noting that MobileNetV2 differs significantly in its network structure from the teacher networks ResUNet and NestedUNet, which further highlights the distinctiveness of our approach.


\section{Experiments}

\subsection{Datasets and Metrics}

\subsubsection{CVPPP}
The CVPPP A1 dataset~\cite{scharr2014annotated} is a well-established plant phenotype dataset, aiming to reveal the relationship between plant phenotypes and genotypes, thus helping to understand the genetic characteristics and genetic mechanisms in biomedical research. The dataset contains images of leaves with complex shapes and significant occlusions and serves as a benchmark dataset for a highly regarded biological instance segmentation task. Each image has a resolution of $530 \times 500$ pixels. In this study, we randomly select 108 images from the dataset for training and 20 images for testing. To evaluate the quality of the segmentation results, we use two widely adopted metrics: symmetric best dice (SBD) and absolute difference in counting ($|$DiC$|$). SBD measures the similarity between the predicted and ground truth segmentation masks, while $|$DiC$|$ counts the absolute difference between the predicted and ground truth number of objects in the image. These metrics are commonly utilized to assess the accuracy of instance segmentation results in computer vision tasks. 

\subsubsection{BBBC039V1} 
The BBBC039V1 dataset~\cite{ljosa2012annotated} consists of 200 Fluorescence Microscopy (FM) images, each with a resolution of $696 \times 520$ pixels. These images capture U2OS cells exhibiting diverse shapes and densities. We follow the official data split, employing 100 images for training, 50 for validation, and the remaining 50 for testing. To quantitatively evaluate the segmentation results, we adopt four widely used metrics for cell segmentation in FM images. 
Aggregated Jaccard Index ($AJI$) \cite{rand1971objective} measures the similarity between the ground truth and predicted segmentation.
Object-level F1 score ($F1$) \cite{chen2017dcan} measures the accuracy of predicted segmentation at the level of individual cells.
Panoptic Quality ($PQ$)~\cite{kirillov2019panoptic} measures the number of correctly segmented instances and the accuracy of the semantic labeling.
The pixel-level Dice score ($Dice$) measures the similarity between the ground truth segmentation and the predicted segmentation at the pixel level.

\subsubsection{C.elegans} 

The C.elegans dataset~\cite{ljosa2012annotated} is a challenging dataset for image analysis with a large number of organisms in each image. C.elegans itself has a slender shape and often appears in complex overlapping poses, making it difficult to accurately segment individual organisms. The dataset consists of 100 grayscale images, each with a resolution of 696 $\times$ 520 pixels. We partition the dataset into 50 training images and 50 test images using a random split, ensuring that both sets represent the complete dataset adequately.  We use the same four metrics as those used in the BBBC039V1 dataset for quantitative evaluation.

\subsubsection{AC3/4}
AC3 and AC4 are two labeled subsets extracted from the mouse somatosensory cortex dataset~\cite{kasthuri2015saturated}, a widely used electron microscope (EM) dataset for 3D instance segmentation of individual neurons in 2D image sequences. These sequences were acquired at a resolution of $3\times3\times29$ nm. The AC3 dataset consists of 256 sequential images, while the AC4 dataset contains 100 sequential images. For evaluating our proposed method, we partition the data as follows: we use the top 80 sections of AC4 for training, the remaining 20 sections for validation, and the top 100 sections of AC3 for testing.
We adopt two widely used metrics to quantitatively evaluate the segmentation results: the variation of Information (VOI) and the adapted rand error (ARAND). VOI~\cite{meilua2003comparing} measures the distance between two segmentation masks, taking into account both the over-merge and over-segmentation errors. ARAND~\cite{rand1971objective} is a variation of the Rand Index that takes into account the uneven distribution of object sizes in EM image segmentation. Note that lower values of these two metrics indicate better segmentation performance.

\subsubsection{CREMI}
The CREMI dataset~\cite{cremi}, which is imaged from adult Drosophila melanogaster brain tissue at a resolution of $4\times4\times40$ nm, is another EM dataset used for 3D neuron segmentation. It is composed of three sub-volumes (CREMI-A/B/C) that correspond to different neuron types, with each sub-volume consisting of 125 consecutive images. Each sub-volume is split into 50 sections for training, 25 sections for validation, and 50 sections for testing. We adopt the same quantitative metrics (VOI and ARAND) as those used for the AC3/4 to evaluate the results on the CREMI dataset.

\subsection{Implementation Details}

Throughout our experiments, we conduct our computations within a well-defined environment comprising PyTorch 1.0.1, CUDA 9.0, and Python 3.7.4. To optimize model training, we utilize the Adam optimizer with $\beta_1 = 0.9$ and $\beta_2 = 0.99$, a learning rate of $10^{-4}$, and a batch size of 2. These choices ensure efficient and effective training processes. We utilize a single NVIDIA TitanXP GPU for training, and conduct 300K iterations for each model. To address GPU memory limitations, we follow~\cite{huang2022learning} to set the embedding dimension of the last layer to 16. Additionally, we compute affinities by considering adjacent pixel embeddings within $N=1$ voxel stride for 3D networks, and within $N=27$ pixel strides for 2D networks. The hyper-parameters $K$ and $L$ of the memory bank mechanism are set as 32 and 12, respectively.

\begin{table*}[!htb]\fontsize{8}{10}\selectfont
\centering
\renewcommand\tabcolsep{1pt}
	\caption{Quantitative comparison of different knowledge distillation methods on 2D biomedical instance segmentation datasets. We conduct experiments on four sets of teacher-student network pairs consisting of two teacher networks and two student networks. A bold score represents the best performance on the corresponding dataset. }
 \label{cvp}
	\begin{center}
        \renewcommand\tabcolsep{8.5pt}
		\begin{tabular}{l|cc|cccc|cccc}
   \toprule[1.2pt]
 \multicolumn{1}{c|}{\multirow{2}{*}{Methods}} & \multicolumn{2}{c|}{CVPPP} & \multicolumn{4}{c|}{C.elegans} & \multicolumn{4}{c}{BBBC039V1
} \\
	\cmidrule(r){2-11}
		 ~ &  SBD $\uparrow$ & $|$DiC$|$ $\downarrow$ & AJI $\uparrow$ & Dice $\uparrow$ & F1 $\uparrow$ & PQ $\uparrow$& AJI $\uparrow$ & Dice $\uparrow$ & F1 $\uparrow$ & PQ $\uparrow$ \\
		\midrule
	    T1: ResUNet     & 88.6 & 1.15    & 0.816 &0.916 &0.931 &0.794 &0.899 &0.956 &0.962 &0.878   \\
        T2: NestedUNet  & 88.4 & 1.10    &0.810 &0.902 &0.928 &0.782  &0.900 &0.957 &0.962 &0.879   \\
        \midrule
        S1: ResUNet-tiny & 81.9 & 2.15 &0.730 &0.837 &0.901 &0.684 &0.865 &0.928 &0.959 &0.840  \\

        S2: MobileNet & 71.9 & 5.00  &0.548 &0.614 &0.812 &0.442 &0.737 &0.905 &0.887 &0.709   \\
        \midrule
        \midrule
        
        T1 \& S1 + AT~\cite{zagoruyko2016paying} &  83.9  & 1.60 &0.749 &0.875 &0.904 &0.731 &0.875 &0.937 &0.959 &0.854 \\
        T1 \& S1 + SPKD~\cite{tung2019similarity}  & 85.2 & 1.30 &0.740 &0.853 &0.904 &0.709 &0.881 &0.940 &\textbf{0.962} &0.858  \\
        T1 \& S1 + ReKD~\cite{chen2021distilling}  & 85.6  & 1.25 & 0.708 & 0.859  &0.883 & 0.723 &0.879 &0.945 &0.961 &0.861 \\
        T1 \& S1 + BISKD~\cite{liu2022efficient} & 86.4 & 1.15 &0.760 &0.865 &0.912 &0.734 &0.872 &0.934 &0.959 &0.851   \\
        T1 \& S1 + Ours  & \textbf{87.0}  & \textbf{1.15} & \textbf{0.765} & \textbf{0.884} & \textbf{0.918} & \textbf{0.759} &\textbf{0.884} &\textbf{0.946} &0.961 &\textbf{0.868}   \\
        \midrule
        \midrule

        T1 \& S2 + AT~\cite{zagoruyko2016paying} &  82.8  & 1.40 &0.545 &0.627 &0.817 &0.451 &0.760 &0.931 &0.893 &0.740   \\
        T1 \& S2 + SPKD~\cite{tung2019similarity}  & 73.8 & 3.95 &0.552 &0.612 &0.819 &0.444  &0.749 &0.917 &0.890 &0.725   \\
        T1 \& S2 + ReKD~\cite{chen2021distilling}  & 81.5  & 1.60 &0.595 &0.732 &0.828 &0.528 &0.757 &0.924 &0.894 &0.737   \\
        T1 \& S2 + BISKD~\cite{liu2022efficient} &84.7 & 1.40 &0.655 &0.799 &0.851 &0.611 & 0.766 &0.930 &0.895 &0.746  \\
        T1 \& S2 + Ours  & \textbf{86.0} & \textbf{1.10} &\textbf{0.672} &\textbf{0.839} &\textbf{0.857} &\textbf{0.645} &\textbf{0.771} &\textbf{0.938} &\textbf{0.896} &\textbf{0.753}   \\
        \midrule
        \midrule
        
        T2 \& S1 + AT~\cite{zagoruyko2016paying}  & 84.7  & 1.25 &0.750 &0.869 &0.908 &0.727 &0.877 &0.940 &0.960 &0.855 \\
        T2 \& S1 + SPKD~\cite{tung2019similarity} & 84.0  & 1.60 &0.739 &0.844 &0.912 &0.702 &0.878 &0.942 &0.959 &0.859  \\
        T2 \& S1 + ReKD~\cite{chen2021distilling} & 85.1  & 1.45 &0.749 &0.861 &0.904 &0.726 &0.880 &0.942 &0.959 &0.857   \\
        T2 \& S1 + BISKD~\cite{liu2022efficient}  & 83.6  & 1.20 &0.702 &0.853 &0.882 &0.713 & 0.883 &0.947 &0.961 &0.865   \\
        T2 \& S1 + Ours  & \textbf{85.8}  & \textbf{1.10} & \textbf{0.751} & \textbf{0.874} & \textbf{0.912} & \textbf{0.746} &\textbf{0.884} &\textbf{0.948} &\textbf{0.963} &\textbf{0.870}   \\
        \midrule
        \midrule
        T2 \& S2 + AT~\cite{zagoruyko2016paying} &  84.4  & \textbf{1.00} &0.549 &0.630 &0.816 &0.454 &0.763 &0.930 &0.892 &0.738   \\
        T2 \& S2 + SPKD~\cite{tung2019similarity}  & 72.9 & 4.60 &0.558 &0.625 &0.822 &0.454 &0.751 &0.912 &0.889 &0.724   \\
        T2 \& S2 + ReKD~\cite{chen2021distilling}  & 79.7  & 2.65 &0.563 &0.652 &0.816 &0.478 &0.750 &0.919 &0.890 &0.727   \\
        T2 \& S2 + BISKD~\cite{liu2022efficient} & 84.9 & 1.25 &0.679 &0.834 &0.864 &0.645 &0.768 &0.933 &\textbf{0.897} &0.749  \\
        T2 \& S2 + Ours  & \textbf{85.3} & 1.15 &\textbf{0.697} &\textbf{0.871} &\textbf{0.867} &\textbf{0.679} &\textbf{0.776} &\textbf{0.939} &0.890 &\textbf{0.757} \\
	\bottomrule[1.2pt]
	\end{tabular}
    \end{center}
\end{table*}

\begin{figure*}[!htb]
\includegraphics[width= \textwidth]{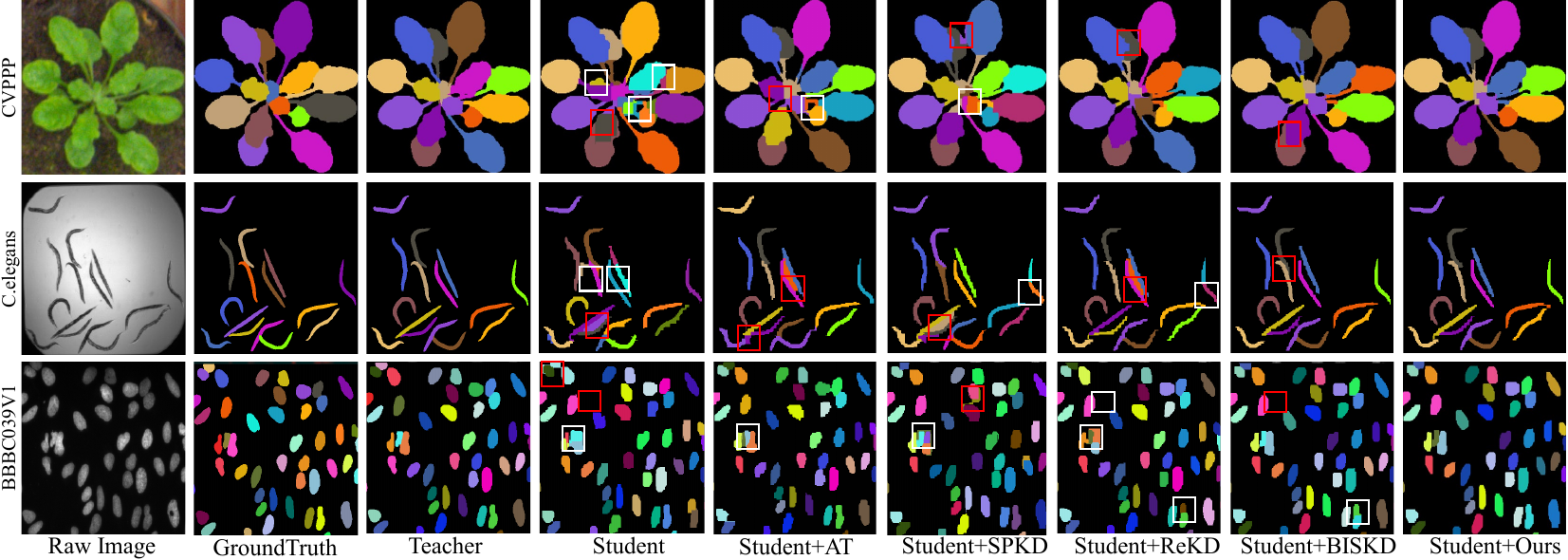}
\caption{Visual comparisons on three 2D datasets.  We use networks ResUNet (T1) and MobileNet (S2) as the teacher and student networks, respectively.
Over-merge and over-segmentation in the results of the student network are highlighted by red and white boxes, respectively.}
\label{2d_total}
\end{figure*}

\begin{table*}[!htb]
\fontsize{7.6}{10}\selectfont
 \caption{Quantitative comparison of different knowledge distillation methods on 3D biomedical instance datasets, where we use the 3D UNet MALA and its corresponding tiny version as the teacher-student network pair. Two post-processing algorithms (waterz~\cite{funke2018large} and LMC~\cite{beier2017multicut}) are adopted to generate final segmentation results. VOI/ARAND are adopted as metrics.}\label{cremi}
 \begin{center}
        \renewcommand\tabcolsep{5.9pt}
 \begin{tabular}{l|c|c|c|c|c|c|c|c} 
        \toprule[1.2pt]
   \multirow{2}{*}{U-Net MALA} & \multicolumn{2}{c|}{AC3/4} & \multicolumn{2}{c|}{CREMI-A} & \multicolumn{2}{c|}{CREMI-B} & \multicolumn{2}{c}{CREMI-C} \\
   \cmidrule{2-9}
   & Waterz $\downarrow$ & LMC $\downarrow$ & Waterz $\downarrow$ & LMC $\downarrow$ & Waterz $\downarrow$ & LMC $\downarrow$ & Waterz $\downarrow$ & LMC $\downarrow$ \\
   \midrule
   T: MALA  & 1.296 / 0.115  & 1.261 / 0.110 & 0.853 / 0.132 & 0.846 / 0.132 & 1.653 / 0.129 & 1.503 / 0.091 & 1.522 / 0.123 & 1.618 / 0.205 \\
   S: MALA-tiny & 1.649 / 0.122 & 1.565 / 0.122  &1.098 / 0.182 & 0.961 / 0.147 & 2.037 / 0.171 & 1.782 / 0.120 & 2.085 / 0.241 & 1.733 / 0.203\\
   \midrule
   AT~\cite{zagoruyko2016paying} & 1.496 / 0.119 & 1.469 / 0.115 & 1.068 / 0.176  & 0.905 / \textbf{0.132} & 1.961 / 0.165 & 1.774 / 0.155 & 1.805 / 0.151  & 1.691 / 0.226   \\
  SPKD~\cite{tung2019similarity} & 1.463 / 0.115 & 1.444 / 0.113 & 0.962 / 0.150  & 0.895 / 0.140 &   1.785 / 0.150  & 1.716 / 0.117 &  1.750 / 0.163  & 1.674 / 0.227  \\
  ReKD~\cite{chen2021distilling} & 1.428 / 0.115 & 1.385 / 0.109 & 0.932 / 0.149  & 0.879 / 0.135 & 1.887 / 0.148  & 1.655 / 0.115 & 1.649 / 0.126  & 1.684 / 0.199  \\
  BISKD~\cite{liu2022efficient}  & 1.384 / 0.120 & 1.334 / 0.116 & 0.892 / 0.139  & 0.856 / 0.136 & 1.739 / 0.140 & 1.598 / \textbf{0.113}  & 1.595 / \textbf{0.119} & 1.567 / 0.159   \\
        Ours  & \textbf{1.320} / \textbf{0.108}  & \textbf{1.279} / \textbf{0.103}  & \textbf{0.853} / \textbf{0.138} & \textbf{0.821} / 0.135  &\textbf{1.524} / \textbf{0.100}  &\textbf{1.542} / 0.127 &  \textbf{1.568} / 0.125 & \textbf{1.470} / \textbf{0.102}   \\
  \bottomrule[1.2pt]
		\end{tabular}
	\end{center}
\end{table*}

\begin{figure*}[!h]
\includegraphics[width= \textwidth]{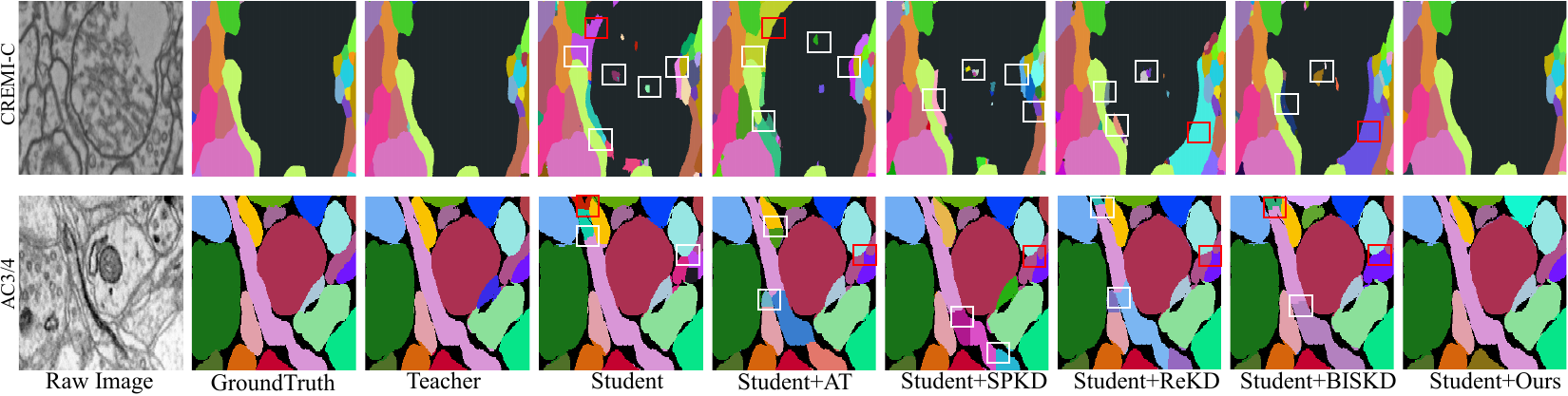}
\caption{2D visual comparisons of segmentation results on the CREMI-C and AC3/4 dataset.} 
\label{2d_ac3}
\end{figure*}

\section{Experimental results}

\subsection{Baseline Methods }
We perform a comparative analysis between our proposed method and three state-of-the-art knowledge distillation methods that are widely used for feature maps, which include:

1) Attention Transferring (AT)~\cite{zagoruyko2016paying}: This method involves the transfer of attention maps from a teacher network to a student network. These attention maps highlight the most relevant regions of the input image for the task at hand, providing guidance to the student network during training.

2) Similarity Preserving Knowledge Distillation (SPKD)~\cite{tung2019similarity}: This method focuses on maintaining similarity between the intermediate feature maps of the teacher and student networks. By minimizing the difference between these feature maps, the student network is encouraged to produce similar results to the teacher network.

3) Review Knowledge Distillation (ReKD)~\cite{chen2021distilling}: This method adopts a novel review mechanism for knowledge distillation, which utilizes the multi-level information from the teacher network to guide the one-level feature learning of the student network. 

4) Biomedical Instance Segmentation Knowledge Distillation (BISKD): This is our preliminary work~\cite{liu2022efficient} tailored for biomedical instance segmentation.

\subsection{Results on 2D Datasets}

\begin{figure*}[!t]
\includegraphics[width= 1 \textwidth]{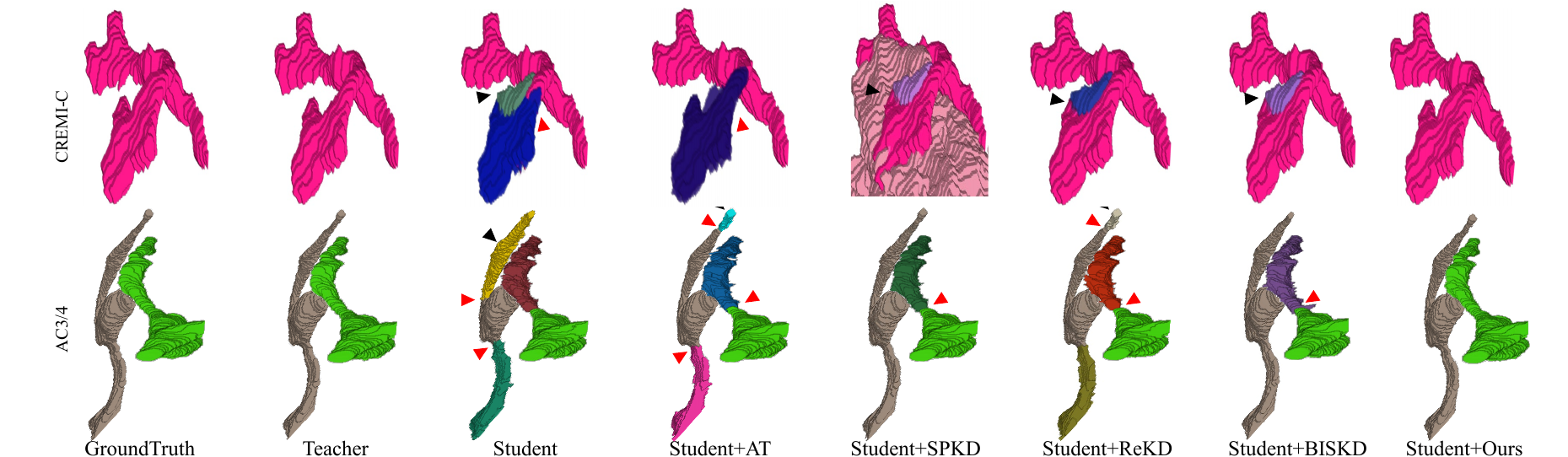}
\caption{3D visual comparisons on the CREMI-C and AC3/4 dataset. Red and black arrows indicate over-segmentation and over-merge, respectively.} 
\label{3d_ac3}
\end{figure*}

We demonstrate the effectiveness of our knowledge distillation method on three 2D biomedical datasets CVPPP, C.elegans, and BBBC039V1. From the results in Table~\ref{cvp}, we can observe that: 

(1) Our proposed method consistently outperforms existing distillation methods and significantly reduces the performance gap between student and teacher networks in various experimental results. Compared to the second best distillation methods, on the CVPPP dataset, the ResUNet-tiny and MobileNet student networks achieve improvements of 5.9\% and 19.6\% for the SBD metric. On the C.elegans dataset, the improvements for the (AJI, Dice, F1, PQ) metrics are (4.8\%, 5.6\%, 1.9\%, 11.0\%) and (27.1\%, 33.1\%, 6.8\%, 53.6\%) respectively. On the BBBC039V1 dataset, the improvements are (2.2\%, 1.1\%, 0.4\%, 3.6\%) and (5.3\%, 3.4\%, 1.0\%, 6.8\%) respectively. Additionally, our method reduces the performance gap between ResUNet and ResUNet-tiny networks by 71.6\% and 84.4\% for the SBD metric on the CVPPP dataset, and (40.7\%, 59.5\%, 56.7\%, 68.2\%) and (46.3\%, 74.5\%, 37.8\%, 57.7\%) for the (AJI, Dice, F1, PQ) metrics on the C.elegans dataset. On the BBBC039V1 dataset, the improvements are (37.5\%, 82.1\%, 66.7\%, 73.7\%) and (21.0\%, 64.7\%, 12.0\%, 26.0\%).

(2) Our knowledge distillation method proves to be highly effective even when dealing with teacher-student network pairs that have significantly different network structures, such as experimental settings with MobileNet as the student network. This highlights the versatility of our method and demonstrates its ability to reduce the performance gap between such teacher and student networks.

(3) Baseline methods AT, SPKD, and ReKD ignore the key knowledge of instance-level features and instance relations, which hinders their ability to guide the student network in enlarging the difference between adjacent instances and reducing the feature variance of pixels within the same instance. This limitation often leads to significant over-merging and over-segmentation. Furthermore, these baseline methods neglect the importance of instance boundary structure knowledge, which leads to additional segmentation errors and coarse boundaries.

(4) Our preliminary work BISKD only focuses on individual input images and neglects inter-image semantic instance relations. This limits the effectiveness of the knowledge transfer process and leads to suboptimal segmentation results.

In addition to the quantitative results, we conduct visual comparisons between the segmentation results of our proposed distillation method and those of the baseline methods on challenging cases, as depicted in Fig.~\ref{2d_total}. These visual comparisons clearly demonstrate the superiority of our distillation method in terms of segmentation performance. Notably, our method exhibits a stronger ability to differentiate adjacent instances and predict more precise instance boundaries, effectively mitigating issues related to over-merging and over-segmentation. This capability is particularly crucial for challenging cases involving objects with complex shapes and distributions, where baseline methods tend to struggle with errors of over-merging and over-segmentation. Our distillation method effectively overcomes these limitations, resulting in more accurate and refined segmentation results.

\subsection{Results on 3D Datasets}

We compare various knowledge distillation methods for the 3D U-Net MALA on the AC3/4 dataset and three CREMI subvolumes, as presented in Table~\ref{cremi}. Our proposed method consistently outperforms the competing distillation methods, exhibiting statistically significant improvements in the majority of experiments. Specifically, our method achieves a remarkable reduction in the performance gap between the student and teacher networks, with reductions exceeding 93.3\% for the key VOI metric on the AC3/4 dataset and over 72.9\% for the VOI metric on the CREMI datasets. The student network demonstrates substantial improvements of 20.0\%, 22.7\%, 25.5\%, and 24.9\% for the VOI metric on the AC3/4, CREMI-A, CREMI-B, and CREMI-C datasets, respectively.

We present the 2D visual comparison in Fig.~\ref{2d_ac3}, showcasing the superior performance of our method in enabling the student network to accurately distinguish instances and address over-segmentation and over-merge errors. Additionally, the 3D visual comparison in Fig.~\ref{3d_ac3} highlights the distinct advantage of our proposed method in preserving the accuracy of neuron structures compared to existing methods.


\subsubsection{Analysis on visualized embeddings}

\begin{figure}[!tb]
\includegraphics[width=0.47 \textwidth]{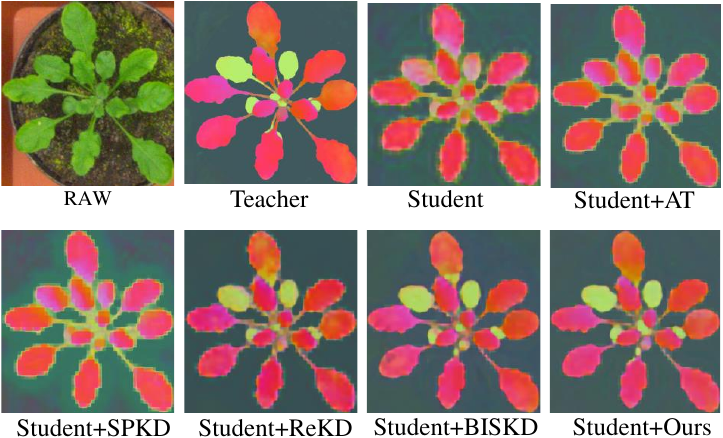}
\caption{A visual example of different embedding maps predicted by student networks distilled with different knowledge distillation methods. We use networks ResUNet (T1) and MobileNet (S2) as the teacher and student networks, respectively.} %
\label{vis_emb}
\end{figure}

To facilitate a comprehensive analysis of the functionality of the proposed knowledge distillation method, we present a visualization of the embeddings generated by the student networks, which have been distilled using various distillation methods. To achieve this, we employ the PCA technique to project the embeddings from a high-dimensional space onto a 3-dimensional RGB color space in Fig. \ref{vis_emb}. Based on the visual results, we have made three observations:

(1) The embeddings predicted by the student model may not adequately capture the relation between adjacent instances, leading to embeddings of neighboring instances having similar RGB color, \textit{i.e.}, similar feature representation. Furthermore, the instance boundary regions in the embedding map appear blur and lack accurate structural information.

(2) When compared to baseline methods, the visualized embeddings obtained from the student network using our knowledge distillation method exhibit more distinct color differences among adjacent instances. Additionally, the instance boundary regions demonstrate clear and accurate structures. These observations indicate that our proposed IGD and AGD schemes effectively facilitate the student network in learning instance relations in the feature space and capturing pixel-level boundary structure information.

(3) In comparison to the visualized embeddings from the `Student+BISKD' approach, the embeddings from the `Student+Ours' method exhibit more pure colors within each instance area. This observation confirms the importance of considering cross-image relations.

\subsection{Ablation Study}

\begin{table}[!t]\fontsize{8.9}{10}\selectfont
\centering
\caption{An ablation study is conducted on the CVPPP dataset to evaluate the performance of different components of our distillation method in the teacher-student network pair of ResUNet and MobileNet networks. The check mark and cross mark indicate the usage and non-usage of this component, respectively.}
\label{tab:ablation_study}
\renewcommand\tabcolsep{4pt}
\begin{tabular}{cccccccc}
\toprule[1.2pt]
$\mathcal{L}_{Node}^{Intra}$ & $\mathcal{L}_{Edge}^{Intra}$ & $\mathcal{L}_{AGD}^{Intra}$ & $\mathcal{L}_{Edge}^{Inter}$ & $\mathcal{L}_{AGD}^{Inter}$ & SBD $\uparrow$ & $|$DiC$|$ $\downarrow$ \\
\midrule
 \XSolidBrush & \XSolidBrush & \XSolidBrush & \XSolidBrush & \XSolidBrush & 71.9 & 5.00\\
\Checkmark & \XSolidBrush & \XSolidBrush & \XSolidBrush & \XSolidBrush &76.5 &2.45 \\
\Checkmark & \Checkmark & \XSolidBrush & \XSolidBrush & \XSolidBrush &79.4 &2.50 \\
\Checkmark & \Checkmark & \Checkmark & \XSolidBrush & \XSolidBrush &84.7 &1.40 \\
\Checkmark & \Checkmark & \Checkmark & \Checkmark & \XSolidBrush &85.2 &1.35 \\
\Checkmark & \Checkmark & \Checkmark & \Checkmark & \Checkmark & \textbf{86.0} & \textbf{1.10} \\
\bottomrule[1.2pt]
\end{tabular}
\end{table}

\begin{table}[t]\fontsize{8.9}{10}\selectfont
\centering
\caption{Ablation study on loss weight hyperparameters on the CVPPP dataset. We adopt the same hyperparameters for all experimental settings and use networks ResUNet (T1) and MobileNet (S2) as the teacher and student networks for analysis.}
\label{tab:hyperparameters}
\renewcommand\tabcolsep{8pt}
\begin{tabular}{ccccccc}
\toprule[1.2pt]
$\lambda_1$ & $\lambda_2$ & $\lambda_3$ & $\lambda_4$ 
 & $\lambda_5$ & SBD $\uparrow$ & $|$DiC$|$ $\downarrow$ \\
\midrule
0.1 & 0.1 & 10 & 1 & 1 & \textbf{86.0} & 1.10  \\
1 & 0.1 & 10 & 1 & 1 & 85.3 & 1.10  \\
0.01 & 0.1 & 10 & 1 & 1 & 85.6 & 1.30  \\
0.1 & 1 & 10 & 1 & 1 & 85.8  & 1.35  \\
0.1 & 0.01 & 10 & 1 & 1 & 85.2 & 1.35  \\
0.1 & 0.1 & 100 & 1 & 1 & 84.9 & 1.20  \\
0.1 & 0.1 & 1 & 1 & 1 & 84.7 & 1.25  \\
0.1 & 0.1 & 10 & 0.1 & 1 & 84.9 & 1.15  \\
0.1 & 0.1 & 10 & 10 & 1 & 84.8 & 1.45  \\
0.1 & 0.1 & 10 & 1 & 0.1 & 85.3 & \textbf{0.80}  \\
0.1 & 0.1 & 10 & 1 & 10 & 84.9 & 1.30  \\
\bottomrule[1.2pt]
\end{tabular}
\end{table}

\subsubsection{Effectiveness of different distillation components}
To verify the effectiveness of the distillation components of our method, we conduct the ablation study on the proposed IGD (divided into edge and node parts) and AGD schemes. These two schemes work at both the intra-image level and the inter-image level. Thus, we validate these loss terms in our distillation method, including  $\mathcal{L}_{Node}^{Intra}$, $\mathcal{L}_{Edge}^{Intra}$, $\mathcal{L}_{AGD}^{Intra}$, $\mathcal{L}_{Edge}^{Inter}$ and $\mathcal{L}_{AGD}^{Inter}$. The results presented in Table~\ref{tab:ablation_study} demonstrate the positive impact of each component on enhancing the performance of the student network and narrowing the performance gap with the teacher network. Notably, the final row of the table highlights that our architecture, incorporating all loss components, achieves the highest performance among the evaluated configurations.

The IGD scheme at the intra-image level, consisting of $\mathcal{L}_{Node}^{Intra}$ and $\mathcal{L}_{Edge}^{Intra}$, improves the undistilled student model by 10.4\% according to the SBD metric. This improvement further increases to 17.8\% when the AGD scheme at the intra-image level is also incorporated. Extending these schemes to the inter-image level boosts the improvement to 19.6\%. The IGD and AGD schemes demonstrate complementary effects, as do the intra-image level and inter-image level schemes.

\begin{figure}[!tb]
\includegraphics[width=0.48 \textwidth]{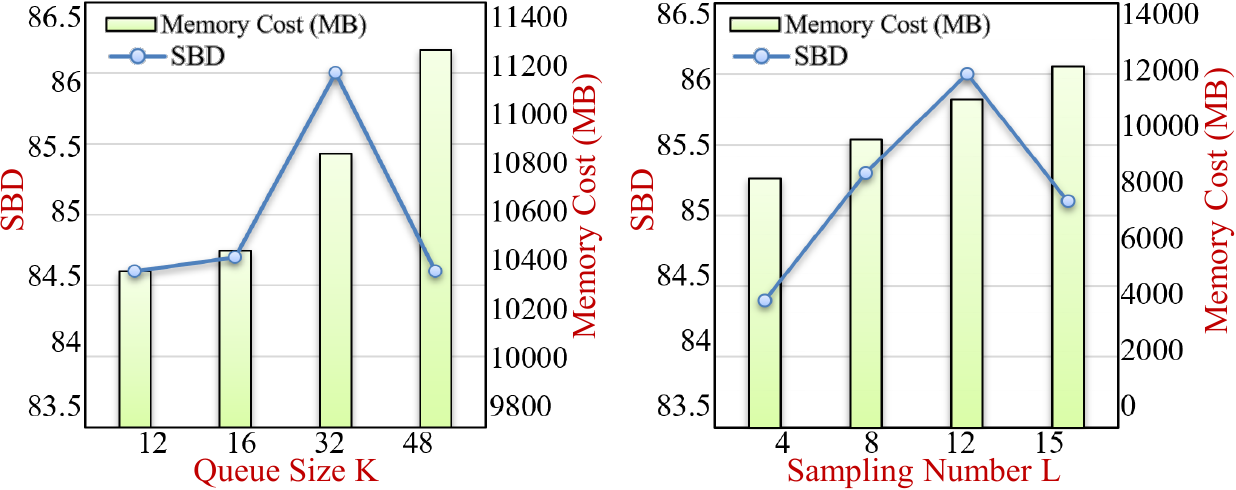}
\caption{Ablation study on the queue size $K$ and sampling number $L$. Experiments are performed for the teacher-student network pair of ResUNet and MobileNet networks on the CVPPP dataset. `Memory Cost' denotes the occupied GPU memory size ($MB$).} %
\label{memory}
\end{figure}

\begin{table}[t]\fontsize{8.9}{10}\selectfont 
\centering
\renewcommand\tabcolsep{3.5pt}
\caption{Ablation study of the distillation performance on a series of small models obtained by reducing the number of channels of each network layer in different ratios. S$_{1/N}$ represents the student networks obtained by reducing the number of channels of the teacher network ResUNet by $\frac{1}{N}$.}
\label{ablation_channel}
	\begin{center}
		\begin{tabular}{l|c|c|c|c}
 			\toprule[1.2pt]
			ResUNet &SBD $\uparrow$ & $|$DiC$|$ $\downarrow$ & \#Params (M) & FLOPs (GMAC) \\
      	    \midrule
   			S$_{1/20}$ w/o KD &74.6 & 3.95 &0.07 & 1.50 \\
   			S$_{1/20}$ w/ KD &\textbf{80.9} &\textbf{2.25} & 0.07 & 1.50  \\
      	    \midrule
			S$_{1/15}$ w/o KD &78.7 &3.10  & 0.17 & 3.28 \\
   			S$_{1/15}$ w/ KD  &\textbf{84.2} &\textbf{1.85}  & 0.17 & 3.28  \\
      	    \midrule
			S$_{1/10}$ w/o KD &81.9 &2.15  & 0.30 & 5.76 \\
   			S$_{1/10}$ w/ KD  &\textbf{87.0} &\textbf{1.15}  & 0.30 & 5.76  \\
      	    \midrule
			S$_{1/5}$ w/o KD &85.1 &1.50 & 0.90 & 17.31 \\
   			S$_{1/5}$ w/ KD &\textbf{87.6} &\textbf{1.25}  & 0.90 & 17.31  \\
            \bottomrule[1.2pt]
		\end{tabular}
	\end{center}
\end{table}

\subsubsection{Sensitivity experiments on hyperparameters}

Given the importance of hyperparameters in the distillation method's loss terms for achieving optimal performance, we conduct a sensitivity analysis on these hyperparameters to effectively balance different loss functions. Table \ref{tab:hyperparameters} presents the results of our analysis.
Based on the findings, it is observed that $\lambda_1$ and $\lambda_2$ have a relatively minor impact on the performance metrics within the considered ranges. However, $\lambda_3$, $\lambda_4$, and $\lambda_5$ emerge as critical hyperparameters for achieving optimal performance.

\subsubsection{Impact of the queue size and sampling number}

We perform an ablation study to examine the effect of the queue size $K$ and sampling number $L$ of the memory bank mechanism on the distillation performance, as shown in Fig.~\ref{memory}. It can be observed that the larger value of these two hyper-parameters, the more GPU memory overhead. The distillation performance improves as the queue size $K$ and sampling number $L$  increase. This can be attributed to the fact that larger queues provide a more diverse and abundant range of features from different input images, which can capture long-range relations more effectively. However, it is also noted that the distillation performance may saturate at a certain memory capacity. This could be due to the fact that increasing the queue size beyond a certain point may not yield any additional benefits to the model. Therefore, it is essential to strike a balance between these two hyper-parameters and the computational and memory resources available to ensure optimal performance. In addition, compared with sampling number $L$, the queue size $K$ has a greater impact on GPU memory occupation.


\subsubsection{Student networks with different widths}
\label{ablation_models}

We conduct an ablation study on reduced-size models to evaluate the effectiveness of distillation. The models are created by reducing the number of channels in each layer of the networks. Specifically, we generate student networks with width reductions of approximately $\frac{1}{20}$, $\frac{1}{15}$, $\frac{1}{10}$, and $\frac{1}{5}$ compared to the original width. The experiments utilize ResUet network pairs on the CVPPP dataset. The results in Tab.~\ref{ablation_channel} demonstrate that our knowledge distillation method improves the performance of all student networks, even when they have very few parameters. However, it is important to note that the effectiveness of knowledge distillation depends on the initial performance gap between the teacher and student networks. When this gap is minimal, achieving significant improvements becomes challenging.

\section{Conclusion}

In this paper, we propose a novel graph relation distillation approach for biomedical instance segmentation that effectively transfers instance-level features, instance relations, and pixel-level boundaries from a heavy teacher network to a lightweight student network, through a unique combination of instance graph distillation and affinity graph distillation schemes. Furthermore, we extend these two schemes beyond the intra-image level to the inter-image level by incorporating a memory bank mechanism, which captures the global relation information across different input images. Experimental results on both 2D and 3D biomedical datasets demonstrate that our method surpasses existing distillation methods and effectively bridges the performance gap between the heavy teacher networks and their corresponding lightweight student networks.

\ifCLASSOPTIONcaptionsoff
\newpage
\fi

\bibliographystyle{IEEEtran}
\bibliography{IEEEabrv,bib,refs}

\end{document}